\documentclass[runningheads]{llncs}
\PassOptionsToPackage{table}{xcolor}
\usepackage[final,year=2026]{accv}
\usepackage{accvabbrv}
\usepackage{graphicx}
\usepackage{booktabs}
\usepackage{siunitx}
\usepackage{adjustbox}
\usepackage{mathtools}
\usepackage{bm}
\usepackage{enumitem}
\usepackage{placeins}
\usepackage{tikz}
\usetikzlibrary{arrows.meta,positioning,fit}

\setlist[itemize]{topsep=3pt,itemsep=1pt,parsep=0pt}

\usepackage{microtype}
\AtBeginDocument{%
  \setlength{\abovedisplayskip}{6pt plus 2pt minus 2pt}%
  \setlength{\belowdisplayskip}{6pt plus 2pt minus 2pt}%
  \setlength{\abovedisplayshortskip}{2pt plus 1pt}%
  \setlength{\belowdisplayshortskip}{4pt plus 1pt minus 1pt}%
}
\makeatletter
\renewcommand\section{\@startsection{section}{1}{\z@}%
  {-11\p@ \@plus 3\p@ \@minus 3\p@}%
  {7\p@ \@plus 2\p@ \@minus 2\p@}%
  {\normalfont\large\bfseries\boldmath
   \rightskip=\z@ \@plus 8em\pretolerance=10000}}
\renewcommand\subsection{\@startsection{subsection}{2}{\z@}%
  {-9\p@ \@plus 3\p@ \@minus 3\p@}%
  {5\p@ \@plus 2\p@ \@minus 2\p@}%
  {\normalfont\normalsize\bfseries\boldmath
   \rightskip=\z@ \@plus 8em\pretolerance=10000}}
\makeatother

\AtBeginEnvironment{thebibliography}{%
  \renewcommand{\small}{\fontsize{9}{10}\selectfont}%
}

\usepackage[breaklinks,colorlinks,citecolor=accvblue]{hyperref}

\title{Tight Learned Event-Inertial Odometry (TLEIO): a Transformer-Based Architecture for End-to-End Motion from Event Streams}
\titlerunning{Tight Learned Event-Inertial Odometry}

\author{Matteo~Cascone\inst{1}\thanks{Equal contributors.} \and
David~Jensen\inst{1}\textsuperscript{\ensuremath{\star}} \and
Alessandro~Pirini\inst{1}\textsuperscript{\ensuremath{\star}} \and
Angelo~Stefanini\inst{1}\textsuperscript{\ensuremath{\star}} \and
Pietro~Venè\inst{1}\textsuperscript{\ensuremath{\star}} \and Marta~Tintoré~Gazulla\inst{2} \and
Javier~Hidalgo~Carriò\inst{2}}
\authorrunning{M. Cascone et al.}
\institute{ETH Zürich \and Google}

\begin{document}
\maketitle

\begin{abstract}
Event cameras provide microsecond temporal resolution, high dynamic range, and
sparse asynchronous measurements, making them compelling for odometry under fast
motion and challenging illumination. Yet, accurate event-based state estimation
remains difficult: existing systems often use events as a complement to frames
or rely on feature tracking, correspondence search, and optimization-heavy
backends, limiting the efficiency promised by event sensing. We introduce
\emph{Tight Learned Event-Inertial Odometry} (TLEIO), a compact filter-based
pipeline that learns short-window relative motion directly from voxelized event streams
and tightly fuses the resulting constraints with IMU propagation. Its learned
front-end, EventsFormer, predicts local displacements and per-axis uncertainty from event voxels in a single forward pass, while a stochastic-cloning EKF fuses these
overlapping predictions with IMU propagation, producing metric trajectories without explicit feature tracks and pose optimization at inference time. The network is trained directly from pose displacements, without requiring event-flow labels. We evaluate TLEIO on the TartanAir/TartanEvent benchmark and on real DAVIS240c sequences. On TartanAir, TLEIO reduces average ATE from 2.26 m to 1.10 m over the IMU-only TLIO baseline, while achieving up to 48.9\% lower runtime than optimization-based event odometry methods.

\end{abstract}

\section{Introduction}
\label{sec:intro}

Visual Odometry (VO) is an essential component in AR/VR and robotics applications,
but standard frame-based pipelines remain limited by conventional cameras.
Motion blur, low-light scenes, and high dynamic range can degrade feature
tracking and dense alignment, even when the downstream estimator is strong.
Event cameras address these limitations at the sensor level by asynchronously
reporting brightness changes with microsecond temporal resolution, high dynamic
range, and low power consumption. These properties make them well suited to fast
motion and challenging illumination, but their sparse and irregular output
makes odometry from events fundamentally different from frame-based VO~\cite{Gallego20pami}.

\begin{figure}[t]
    \centering
    \includegraphics[width=\textwidth]{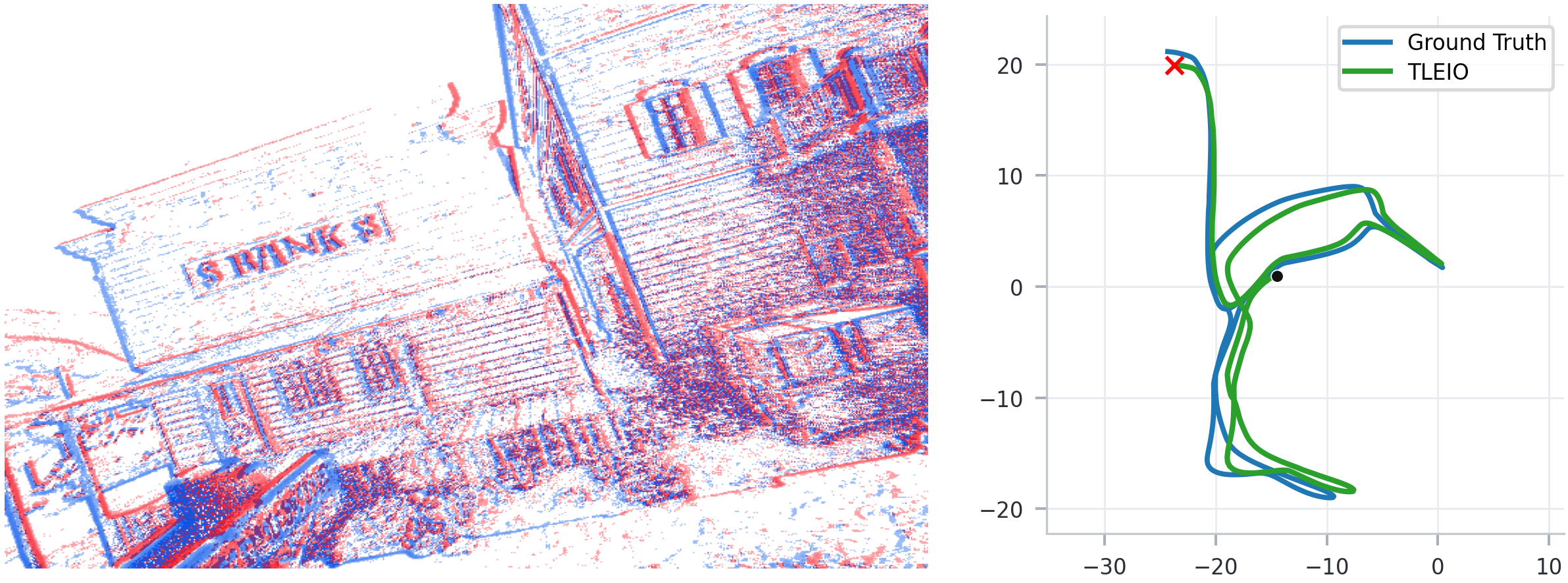}
    \caption{Raw event stream and TLEIO trajectory estimate with ground truth on TartanAir. The black dot represents the start of the trajectory and the red cross the end.}
    \label{fig:eye_catcher}
\end{figure}

Event cameras improve sensing, but they do not by themselves solve odometry.
Since events are sparse, asynchronous, and mainly triggered by motion,
extracting stable geometric correspondences can be difficult, especially in
noisy or uneven event streams. IMUs provide complementary metric short-term
motion information, but they drift when integrated over time. Event-inertial
odometry (EIO) must therefore extract useful motion cues from events and fuse
them with inertial propagation in a stable sequential manner.

Early EIO systems adapted geometric VO ideas to
events. Zhu et al.~\cite{Zhu_2017_CVPR} proposed one of the first EIO pipelines
to estimate 6DoF poses using event-based feature tracking and an Extended Kalman
Filter. Subsequent systems further adapted classical visual-odometry components
to events, while the strongest pipelines often combine frames, events, and IMU
measurements~\cite{Mahlknecht_2022_RAL,Vidal_2018_RAL}. Although effective, these methods
typically rely on explicit correspondences, multi-stage front-ends, or
optimization-heavy back-ends.

We propose Tight Learned Event-Inertial Odometry (TLEIO), which combines a learned event-based front-end, called \textbf{EventsFormer}, with inertial filtering. EventsFormer directly estimates short-window body-frame translations from event voxels. These estimates are then fused with high-rate IMU propagation in a Stochastic Cloning Extended Kalman Filter (SC-EKF) to recover the full 6DoF trajectory. To the best of our knowledge, TLEIO is the first event--inertial odometry framework to learn motion constraints directly from event streams and fuse them with IMU measurements, without explicit feature tracking, optical-flow estimation, or iterative pose optimization. Figure~\ref{fig:eye_catcher} shows an example event stream and the resulting trajectory estimate.

We summarize our contributions as follows:
\begin{itemize}
    \item \textbf{EventsFormer}, a learned event-based front-end that estimates relative translation and per-axis uncertainty from event voxels using divided space--time attention and 2D rotary positional embeddings.
    
 \item \textbf{TLEIO}, which fuses learned motion constraints with IMU propagation in an SC-EKF for full 6DoF estimation without iterative pose optimization, reducing end-to-end runtime by up to 48.9\% compared with optimization-based event-odometry pipelines.
\end{itemize}

\section{Related Work}
\label{sec:related}

\textbf{Event-based odometry.}
Event cameras provide asynchronous measurements at high temporal resolution and
have therefore been widely explored for visual and visual--inertial odometry
~\cite{Lichtsteiner08ssc}. Early approaches relied mainly on geometric cues,
combining event features with filtering or optimization-based state estimation
~\cite{Zhu_2017_CVPR,Weikersdorfer14icra}. Later systems such as EKLT-VIO
~\cite{Mahlknecht_2022_RAL} and Ultimate SLAM~\cite{Vidal_2018_RAL} improved
robustness by combining events with frames and IMU measurements or by using
model-based event tracking. 

\textbf{Learned event-based odometry.}
Recent methods instead learn motion information directly from event
representations. DEVO~\cite{Klenk_2024_3DV} extends learned patch-based odometry
to event streams, DEIO~\cite{Guan_2025_ICCVW} incorporates inertial
measurements, and RAMP-VO~\cite{Pellerito_2024_IROS} combines events and
frames within a learned visual-odometry pipeline. These
approaches achieve strong accuracy, but retain correspondence estimation and
iterative optimization in the odometry backend.

\textbf{Learned filtering and temporal modeling.}
The use of learned local motion estimates within filtering has previously been
shown for inertial odometry. TLIO~\cite{Liu_2020_RAL}, building on stochastic
cloning and MSCKF-style filtering~\cite{4209642}, fuses learned displacement
estimates with IMU propagation in an EKF. TLEIO adopts this filtering principle
for event-based motion estimates. For the learned front-end, EventsFormer builds
on divided space--time attention introduced by TimeSformer~\cite{TimeSformer}
and later used for monocular odometry in TSformer-VO~\cite{TSformerVO}. Here,
the transformer operates on event voxels and predicts relative displacement and
uncertainty, which are directly fused by the SC-EKF.

\section{Methodology}
\label{sec:methodology}

\begin{figure}[t]
  \centering
  \includegraphics[width=\textwidth]{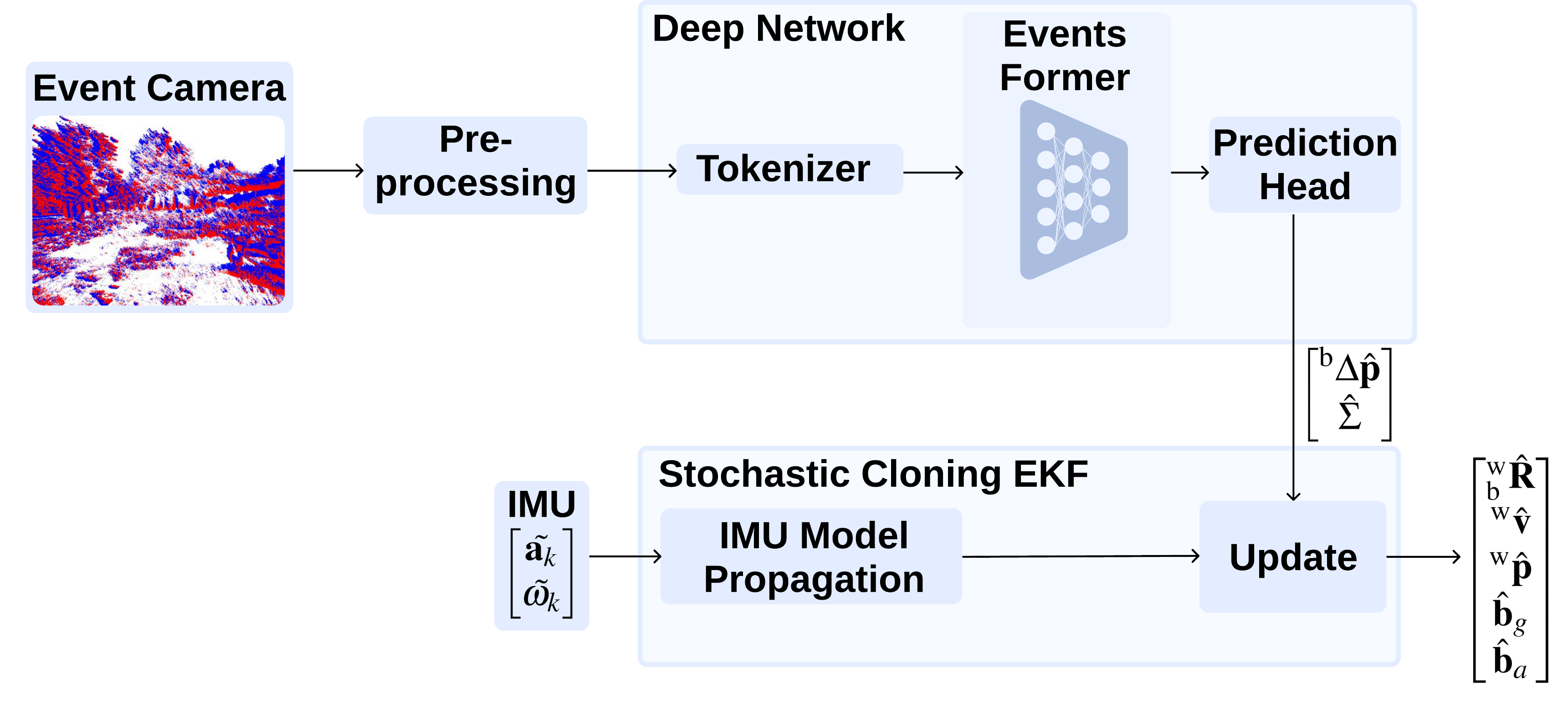}
  \caption{Overview of TLEIO. EventsFormer predicts relative displacement
  ${}^\text{b}\Delta\hat{\mathbf p}$ and diagonal covariance
  $\hat{\boldsymbol\Sigma}$, which are fused with IMU propagation in the SC-EKF
  to estimate the 6DoF pose.}
  \label{fig:model_arch}
\end{figure}

Figure~\ref{fig:model_arch} gives an overview of TLEIO. Raw events are first
converted into voxel grids   (Sec.~\ref{sec:preprocessing})~\cite{zhu_voxels} .
EventsFormer then predicts short-window body-frame 3DoF displacements together
with per-axis uncertainty from voxel clips (Sec.~\ref{sec:network}). These
relative motion estimates are finally fused with high-rate IMU propagation in a
SC-EKF to recover the full 6DoF trajectory (Sec.~\ref{sec:filter}).

\subsection{Pre-processing}
\label{sec:preprocessing}

\paragraph{Event representation.}
Each event $e_k=(t_k,\mathbf u_k,p_k)$ contains its timestamp $t_k$, pixel
location $\mathbf u_k=(x_k,y_k)^T$, and polarity $p_k\in\{-1,+1\}$. Events
within the interval $[t_i-\Delta t,t_i)$ are accumulated into a voxel grid
$\mathbf E_i\in\mathbb R^{C\times H\times W}$ with $C$ temporal bins and sensor
resolution $H\times W$~\cite{zhu_voxels}. Before voxelization, the events are
spatially down-sampled by $d_s$ and isolated events are removed with a local
spatio-temporal background-activity filter~\cite{background_filter}.

\subsection{Network}
\label{sec:network}

EventsFormer builds on the divided space--time attention of
TimeSformer~\cite{TimeSformer}, later used for frame-based visual odometry in
TSformer-VO~\cite{TSformerVO}. We adapt this architecture to event data. The
network takes event voxels as input, uses 2D rotary positional embeddings
(RoPE2D)~\cite{heo2024ropevit,su2023roformer} for spatial encoding, and predicts relative displacement together with
its uncertainty. These predictions are later used as constraints in the SC-EKF.

At reference time $t_i$, the input clip contains the latest $T$ voxels:
\begin{equation}
 \mathcal C_i=\{\mathbf E'_{i-T+1},\ldots,\mathbf E'_i\}
 \in\mathbb R^{T\times C\times H\times W}.
\end{equation}

\begin{figure}[t]
  \centering
  \includegraphics[width=.76\textwidth]{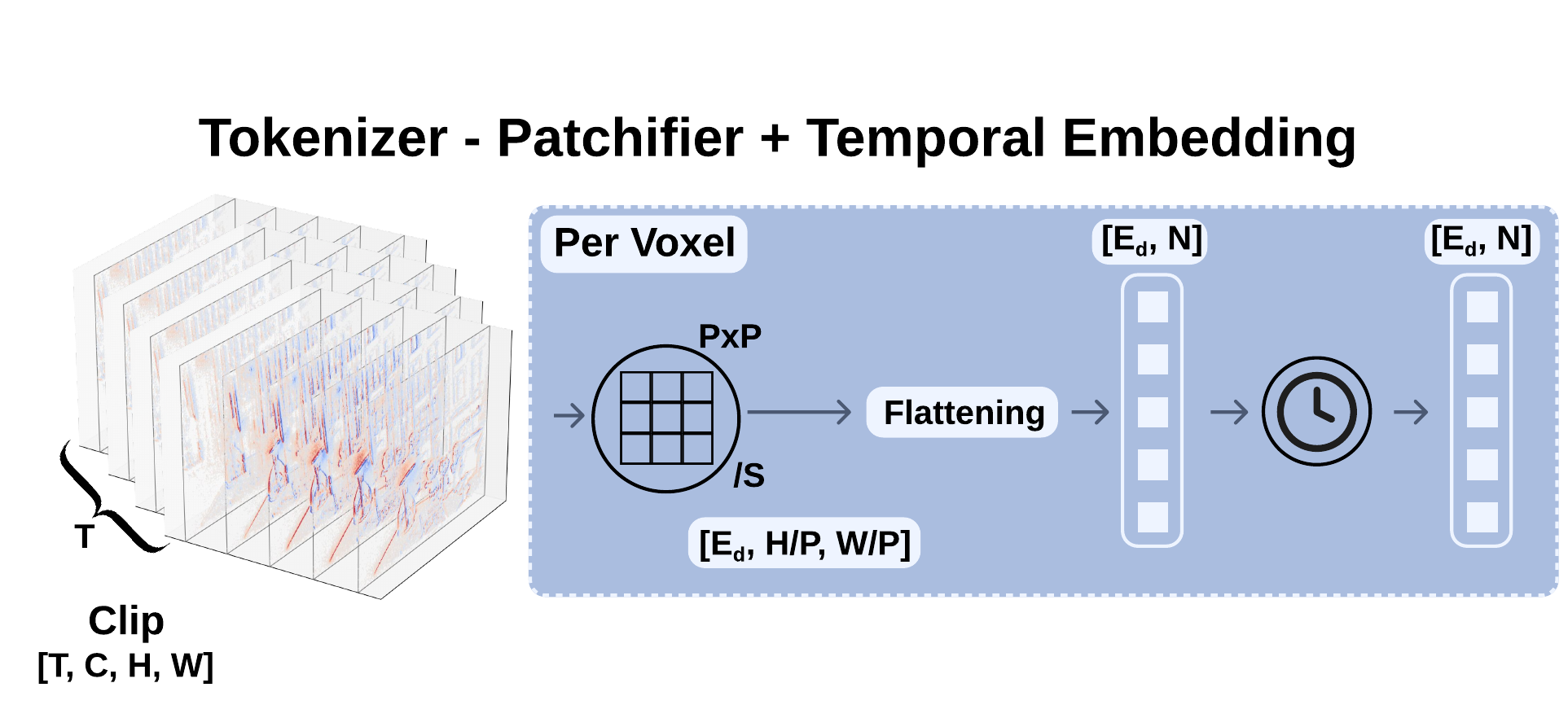}
  \caption{Each event voxel is divided into patches and projected into tokens.
  A temporal embedding is then added to encode the voxel index.}
  \label{fig:tokenizer}
\end{figure}

Each voxel is divided into
$N=\lfloor H/P\rfloor\lfloor W/P\rfloor$ non-overlapping $P\times P$ patches.
A flattened patch $\mathbf x_{(s,t)}\in\mathbb R^{CP^2}$ is projected to the
embedding dimension $E_d$ using a convolution with kernel and stride $S=P$:
\begin{equation}
 \mathbf z^0_{(s,t)}=\mathbf W_p\mathbf x_{(s,t)}+\mathbf b_p.
\end{equation}
Figure~\ref{fig:tokenizer} shows the tokenization process.

The encoder first applies temporal attention across voxels at each spatial
location, followed by spatial attention within each voxel. Unlike TimeSformer
and TSformer-VO, we use RoPE2D instead of learned spatial positional embeddings,
applying it to the queries and keys of spatial attention. A learned temporal embedding is used to
distinguish the voxels in the clip. As in TSformer-VO, a classification token
summarizes the clip. The prediction head outputs $T-1$ body-frame 3DoF
displacements and their per-axis log standard deviations, rather than camera
poses. For encoder layer $\ell$, the updates are

\begin{equation}
\begin{aligned}
 \tilde{\mathbf z}^{\ell}_{s,1:T}
 &=\mathbf z^{\ell-1}_{s,1:T}+\mathrm{FC}\!\left(
 \mathrm{MHSA}_{t}(\mathrm{LN}(\mathbf z^{\ell-1}_{s,1:T}))\right),\\
 \hat{\mathbf z}^{\ell}_{1:N,t}
 &=\tilde{\mathbf z}^{\ell}_{1:N,t}+\mathrm{MHSA}_{s}^{\mathrm{RoPE2D}}\!\left(
 \mathrm{LN}(\tilde{\mathbf z}^{\ell}_{1:N,t})\right),\\
 \mathbf z^{\ell}_{1:N,1:T}
 &=\hat{\mathbf z}^{\ell}_{1:N,1:T}+\mathrm{MLP}\!\left(
 \mathrm{LN}(\hat{\mathbf z}^{\ell}_{1:N,1:T})\right).
\end{aligned}
\end{equation}

\begin{figure}[t]
  \centering
  \makebox[\textwidth][c]{%
    \includegraphics[width=\textwidth]{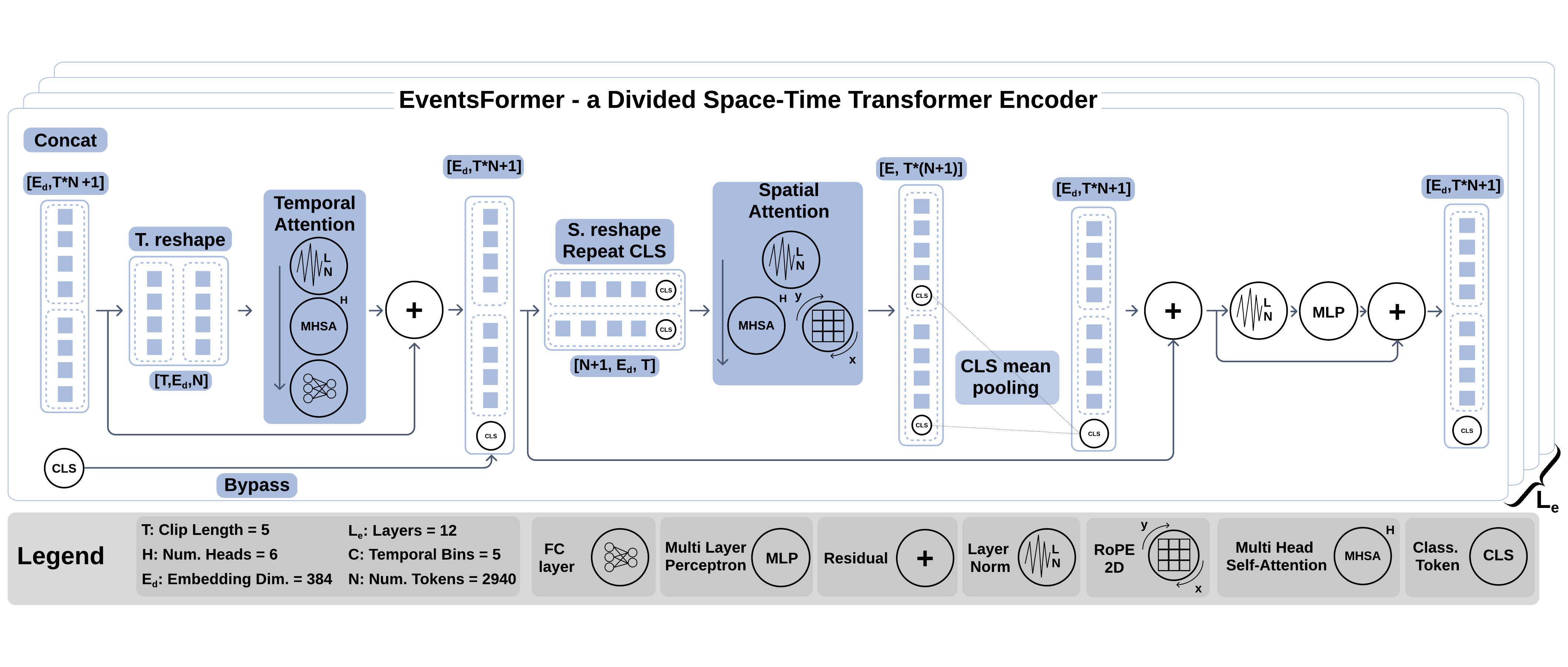}
  }
  \caption{EventsFormer encoder. Temporal and spatial attention are applied in
  sequence, with RoPE2D used in spatial attention. The block is repeated
  $L_e$ times. The CLS token bypasses temporal attention and is averaged across
  voxels after spatial attention.}
  \label{fig:encoder}
\end{figure}

We use $\Delta t=50$ ms, $d_s=0.7$, $C=5$, $P=16$, $E_d=384$, $L_e=12$,
and $T=5$. Compared with the $T=3$ setting used in TSformer-VO, the longer clip
provides more temporal context for motion estimation.
Figure~\ref{fig:encoder} shows the full encoder.

Inference uses a sliding window, so consecutive clips produce overlapping
displacement estimates. Adjacent clips share $T-2$ displacement intervals,
clips two steps apart share $T-3$, and so on. The SC-EKF keeps the corresponding
pose clones and applies each prediction between the appropriate pair. As a
result, the same pose can be updated by several overlapping predictions while it
remains in the window. The predicted standard deviations define the diagonal
covariance of each constraint (Sec.~\ref{sec:training}). The filter preserves
cross-covariances between poses, but does not explicitly model correlations
between overlapping network predictions; the diagonal measurement model is
therefore an approximation.

\subsection{Stochastic-Cloning EKF}
\label{sec:filter}

Following the TLIO implementation~\cite{Liu_2020_RAL}, we use the MSCKF framework~\cite{4209642} to fuse the motion estimates from
EventsFormer with high-rate IMU propagation. The filter maintains four past
poses together with the current state, and uses four consecutive 3D displacement
predictions to form a 12D measurement update. 
\textbf{State.}
The filter state is
$\chi=\{\zeta_1,\ldots,\zeta_m,s\}$, where
$\zeta_j=\{{}_\text{b}^\text{w}\hat{\mathbf R}_j,
{}^\text{w}\hat{\mathbf p}_j\}$ denotes the $j$-th cloned pose and
\[
s=\{{}_\text{b}^\text{w}\hat{\mathbf R},{}^\text{w}\hat{\mathbf v},
{}^\text{w}\hat{\mathbf p},\hat{\mathbf b}_g,\hat{\mathbf b}_a\}
\]
is the current IMU state. We keep $m=4$ clones during propagation and $m=5$
during an update. Using a minimal representation for rotation, the error state is
\begin{equation}
 \delta\chi=\{\delta\zeta_{1:m},\delta s\},\quad
 \delta\zeta_i=\{\delta\boldsymbol\theta_i,\delta\mathbf p_i\},\quad
 \delta s=\{\delta\boldsymbol\theta,\delta\mathbf v,\delta\mathbf p,
 \delta\mathbf b_g,\delta\mathbf b_a\},
\end{equation}
where $\delta$ denotes the difference operator for variables in $\mathbb R^3$,
and the rotation error is defined as
$\delta\boldsymbol\theta \coloneqq\log_{SO(3)}(\mathbf R\hat{\mathbf R}^{-1})$.

\textbf{Propagation.}
Gyroscope and accelerometer measurements are modeled as
$\tilde{\boldsymbol\omega}=\boldsymbol\omega+\mathbf b_g+\mathbf n_g$ and
$\tilde{\mathbf a}=\mathbf a+\mathbf b_a+\mathbf n_a$, with random-walk
biases, according to the statistical IMU model. For an IMU step $\Delta t$, the state is propagated as
\begin{equation}
\begin{aligned}
 {}_\text{b}^\text{w}\hat{\mathbf R}_{k+1}
 &= {}_\text{b}^\text{w}\hat{\mathbf R}_{k}
 \exp_{SO(3)}((\tilde{\boldsymbol\omega}_k-\hat{\mathbf b}_{g,k})\Delta t),\\
 {}^\text{w}\hat{\mathbf v}_{k+1}
 &= {}^\text{w}\hat{\mathbf v}_{k}+{}^\text{w}\mathbf g\Delta t
 +{}_\text{b}^\text{w}\hat{\mathbf R}_{k}
 (\tilde{\mathbf a}_k-\hat{\mathbf b}_{a,k})\Delta t,\\
 {}^\text{w}\hat{\mathbf p}_{k+1}
 &= {}^\text{w}\hat{\mathbf p}_{k}+{}^\text{w}\hat{\mathbf v}_{k}\Delta t
 +\tfrac12\Delta t^2\!\left({}^\text{w}\mathbf g
 +{}_\text{b}^\text{w}\hat{\mathbf R}_{k}
 (\tilde{\mathbf a}_k-\hat{\mathbf b}_{a,k})\right),\\
 \hat{\mathbf b}_{g,k+1}&=\hat{\mathbf b}_{g,k},\qquad
 \hat{\mathbf b}_{a,k+1}=\hat{\mathbf b}_{a,k}.
\end{aligned}
\end{equation}
Rotation is integrated using a third-order quaternion integrator. The linearized
error dynamics and covariance propagation are
\begin{equation}
  \delta s_{k+1}=\boldsymbol\Phi_k^s\delta s_k+\mathbf G_k^s\mathbf n,
  \quad
  \mathbf P_{k+1}=\boldsymbol\Phi_k\mathbf P_k\boldsymbol\Phi_k^\top
  +\mathbf G_k\mathbf Q_k\mathbf G_k^\top,
\end{equation}
where $\mathbf n=[\mathbf n_a,\mathbf n_g,\mathbf n_{b,a},\mathbf n_{b,g}]^T$,
$\mathbf Q_k$ is the process-noise covariance,
$\boldsymbol\Phi_k=\operatorname{diag}(\mathbf I_{6m},$ $\boldsymbol\Phi_k^s)$,
and $\mathbf G_k=[\mathbf 0^T,(\mathbf G_k^s)^T]^T$.
At each measurement time, the current pose is cloned together with its
cross-covariances. After the update, the oldest clone is marginalized.

\textbf{Update.}
For a clone pair $(i,j)$, the relative displacement predicted by the filter is
\begin{equation}
 \mathbf h_{i,j}(\zeta)={}_\text{b}^\text{w}\hat{\mathbf R}_i^T
 ({}^\text{w}\hat{\mathbf p}_j-{}^\text{w}\hat{\mathbf p}_i),
\end{equation}
with residual
$\mathbf r_{i,j}={}^\text{b}\Delta\hat{\mathbf p}_{i,j}-\mathbf h_{i,j}$.
Stacking four such residuals gives $\mathbf r\in\mathbb R^{12}$.
The nonzero blocks of the residual Jacobian are
\begin{equation}
\begin{aligned}
 \mathbf H_{\theta_i}&=-[\mathbf h_{i,j}]_\times, \quad
 \mathbf H_{\delta\mathbf p_i}={}_\text{b}^\text{w}\hat{\mathbf R}_i^T, \quad
 \mathbf H_{\delta\mathbf p_j}=-{}_\text{b}^\text{w}\hat{\mathbf R}_i^T.
\end{aligned}
\end{equation}

Let $\mathbf S$ denote the block-diagonal covariance obtained from the four
network predictions (Sec.~\ref{sec:training}). The Joseph-form EKF update is
\begin{align}
 \mathbf K&=\mathbf P\mathbf H^T
 (\mathbf H\mathbf P\mathbf H^T+\mathbf S)^{-1},\\
 \chi&=\chi\oplus(-\mathbf K\mathbf r),\nonumber\\
 \mathbf P&=(\mathbf I-\mathbf K\mathbf H)\mathbf P
 (\mathbf I-\mathbf K\mathbf H)^T+\mathbf K\mathbf S\mathbf K^T.
\end{align}
The negative sign follows from our residual convention. The operator $\oplus$
applies the error-state correction, with rotations updated as
$\mathbf R'=\mathbf R\exp_{SO(3)}(\delta\boldsymbol\theta)$.
We reject outliers using Mahalanobis gating with the 95th-percentile
$\chi^2$ threshold for 12DoF.

\textbf{Adaptive measurement covariance.}
To reduce the influence of inconsistent network predictions, we use Adaptive
Covariance Estimation~\cite{5462839} to inflate $\mathbf S$ by adding $\mathbf S_k^{\mathrm{infl}}$ when the observed
residuals are larger than expected. Over a window of $M_1$ updates, the empirical
residual covariance is
\begin{equation}
 \mathbf C_k=\frac1{M_1}\sum_{l=k-M_1+1}^{k}\mathbf r_l\mathbf r_l^T
 =\sum_{j=1}^{12}\lambda_{j,k}\mathbf v_{j,k}\mathbf v_{j,k}^T.
\end{equation}
For each eigenvector, the covariance predicted by the filter is
\[
\mu_{j,k}=\mathbf v_{j,k}^T
(\mathbf H\mathbf P\mathbf H^T+\mathbf S)\mathbf v_{j,k}.
\]
If $\max_j(\lambda_{j,k}-\mu_{j,k})>\gamma$, the positive mismatch is added to
the measurement covariance:
\begin{equation}
 \mathbf S_k^{\mathrm{infl}}=\sum_{j=1}^{12}
 \max(\lambda_{j,k}-\mu_{j,k},0)
 \mathbf v_{j,k}\mathbf v_{j,k}^T.
\end{equation}
This reduces the filter's confidence in EventsFormer along directions where the
predicted uncertainty is inconsistent with the observed residuals. The
inflation is removed once
$\max_j(\lambda_{j,k}-\mu_{j,k})<\gamma$ for more than $M_2$ consecutive
updates. 

\subsection{Training}
\label{sec:training}
Following TLIO~\cite{Liu_2020_RAL}, training begins with mean squared error,
\begin{equation}
    \mathcal{L}_{MSE}({}^\text{b}\Delta \hat{\mathbf{p}}, {}^\text{b}\Delta \mathbf{p}) = \frac{1}{n} \sum^n_{i=1}\lVert  {}^\text{b}\Delta \hat{\mathbf{p}}_i - {}^\text{b}\Delta \mathbf{p}_i\Vert_2^2,
\end{equation}
where ${}^\text{b}\Delta\mathbf p_i$ and
${}^\text{b}\Delta\hat{\mathbf p}_i$ are the target and predicted 3D
body-frame displacements. This stabilizes the displacement estimate before
uncertainty learning.

We then use maximum likelihood to learn uncertainty. The network regresses log
standard deviations $\hat{\boldsymbol\ell}_i$, converted to
$\hat{\boldsymbol\sigma}_i=\exp(\hat{\boldsymbol\ell}_i)$ and covariances  $\hat{\boldsymbol\Sigma}_i$:
\begin{equation}
\begin{aligned}
    \mathcal{L}_{ML}({}^\text{b}\Delta \mathbf{p}, \hat{\boldsymbol{\Sigma}}, {}^\text{b}\Delta \hat{\mathbf{p}})
    = \frac{1}{n} \sum^n_{i=1}\bigg(&\frac{1}{2}\log\det(\hat{\boldsymbol{\Sigma}}_i) \\
    &+ \frac{1}{2}\lVert {}^\text{b}\Delta \hat{\mathbf{p}}_i -{}^\text{b}\Delta \mathbf{p}_i\rVert_{\hat{\boldsymbol{\Sigma}}_i^{-1}}^2\bigg) + \mathrm{const.}
\end{aligned}
\end{equation}
where
\begin{equation}
    \hat{\boldsymbol{\Sigma}}_i
    = \mathrm{diag}(\hat{\sigma}_{xi}^2,\hat{\sigma}_{yi}^2,\hat{\sigma}_{zi}^2)
    = \mathrm{diag}(e^{2\hat{\ell}_{xi}},e^{2\hat{\ell}_{yi}},e^{2\hat{\ell}_{zi}})
\end{equation}
for $\hat{\boldsymbol{\ell}}_i=(\hat{\ell}_{xi},\hat{\ell}_{yi},\hat{\ell}_{zi})$.

We train on TartanEvent~\cite{Pellerito_2024_IROS,wang2020tartanairdatasetpushlimits}, which contains 3.5 hours across varied environments, using Distributed Data Parallel on eight NVIDIA A100, 80 GB GPUs for 117 GPU hours. We use a batch size of 32, an initial learning rate of $10^{-4}$, 12 layers and 6 attention heads. The first 60 epochs use MSE loss, while the final 10 use ML loss.

\section{Experiments}
\label{sec:experiments}

We evaluate TLEIO on two complementary benchmarks: the real DAVIS240c dataset
~\cite{Mueggler_2017_IJRR} and the official 16-sequence TartanAir Visual SLAM
test split with corresponding TartanEvent streams
~\cite{Pellerito_2024_IROS,wang2020tartanairdatasetpushlimits}. TartanAir spans
diverse scenes, lighting, and weather conditions, with IMU data  generated at $200\,\mathrm{Hz}$, with gravity $9.80665\,\mathrm{m/s^2}$, Gaussian noise ($3.5\times10^{-3}\,\mathrm{rad/s}$, $0.04\,\mathrm{m/s^2}$), and bias random walks ($5\times10^{-4}\,\mathrm{rad/s/\sqrt{s}}$, $5\times10^{-3}\,\mathrm{m/s^2/\sqrt{s}}$).

\textbf{Metrics.}
We follow the protocols of the respective public benchmarks. On TartanAir, we
align trajectories to ground truth and report Absolute Trajectory
Error (ATE):
\begin{equation}
\mathrm{ATE} =
\sqrt{
\frac{1}{n}
\sum_{i=1}^{n}
\left\lVert {}^{w}\mathbf{p}_i -
{}^{w}\hat{\mathbf{p}}^{\,\mathrm{al}}_i \right\rVert^2
}.
\end{equation}
In particular, we use SE(3) alignment for pipelines that rely on IMU, while Sim(3) is used for the remaining ones with results taken from \cite{Pellerito_2024_IROS}. 
On DAVIS240c, we report Mean Position Error (MPE), normalized by the
ground-truth path length $L$:
\begin{equation}
\mathrm{MPE} =
\frac{100}{Ln}
\sum_{i=1}^{n}
\left\lVert {}^{w}\mathbf{p}_i -
{}^{w}\hat{\mathbf{p}}^{\,\mathrm{al}}_i \right\rVert .
\end{equation}

\textbf{Baselines.}
We compare TLEIO across sensing configurations and estimator designs:
frame-based ORB-SLAM3, COLMAP, DROID-VO, and DPVO; event--frame RAMP-VO;
event--frame--inertial EKLT-VIO; and IMU-only TLIO%
~\cite{Mahlknecht_2022_RAL,Pellerito_2024_IROS,Liu_2020_RAL,
Campos_2021_TRO,Schonberger_2016_CVPR,Teed_2021_DROID,
NEURIPS2023_7ac484b0}. TLIO is trained for 100 epochs on the same split and uses the same
filtering framework, directly isolating the contribution of event-derived
motion constraints. On DAVIS240c, we additionally include event-only DEVO%
~\cite{Klenk_2024_3DV} and event-inertial Ultimate-SLAM and DEIO%
~\cite{Vidal_2018_RAL,Guan_2025_ICCVW}. DEVO and DEIO are excluded from
TartanAir because their released models were trained on the competition test
sequences. RAMP-VO, DEVO, and DEIO also form the runtime comparison as the
closest learned event-based alternatives.

\subsection{Results on the TartanAir Competition Split}
\label{sec:results_tartanair}

Table~\ref{tab:tartanair_competition} reports the ATE over the
16 competition sequences. TLEIO achieves an average error of $1.10$\,m,
reducing TLIO's $2.26$\,m by $51.3\%$ and improving its result on 12
sequences. It also achieves lower average ATE than EKLT-VIO, ORB-SLAM3, and
COLMAP. RAMP-VO ($0.17$\,m), DPVO ($0.21$\,m), and DROID-VO ($0.58$\,m)
remain more accurate. This difference is expected: RAMP-VO combines events
with image frames, while DPVO and DROID-VO are frame-based, and all three
retain visual correspondences across multiple views and iteratively refine
their pose estimates. TLEIO instead predicts each local translation directly
from an event window and fuses it with IMU propagation in a fixed-size
SC-EKF, without correspondence tracking or iterative pose optimization.

\begin{figure}[!htbp]
\centering
\begin{minipage}{0.88\textwidth}
\centering
\parbox[c][1.5em][c]{\linewidth}{\centering\small\textbf{(a) MH02}}\par
\begin{minipage}[c]{.65\linewidth}
\centering
\includegraphics[width=\linewidth]{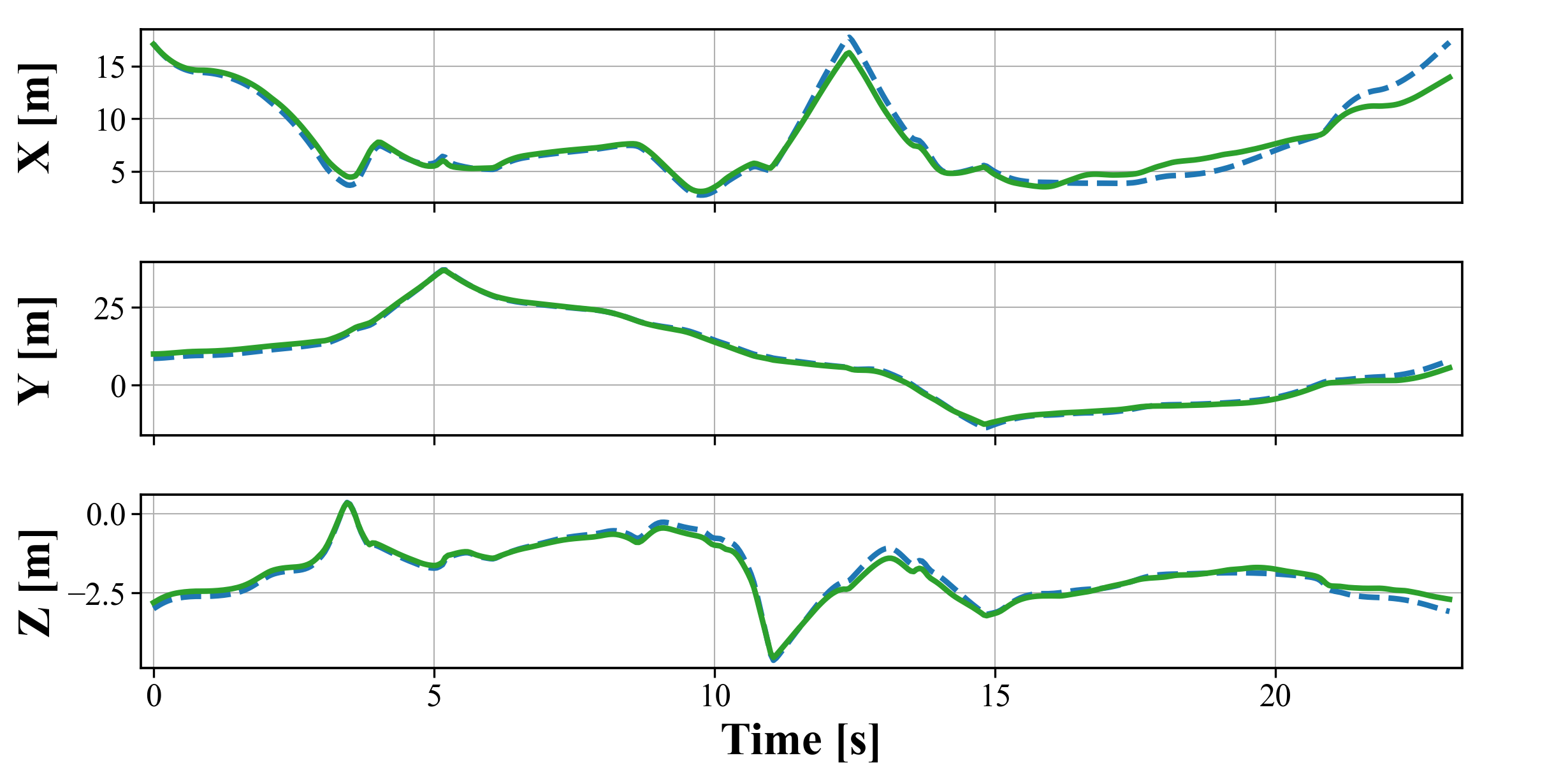}
\end{minipage}\hfill
\begin{minipage}[c]{.325\linewidth}
\centering
\includegraphics[width=\linewidth]{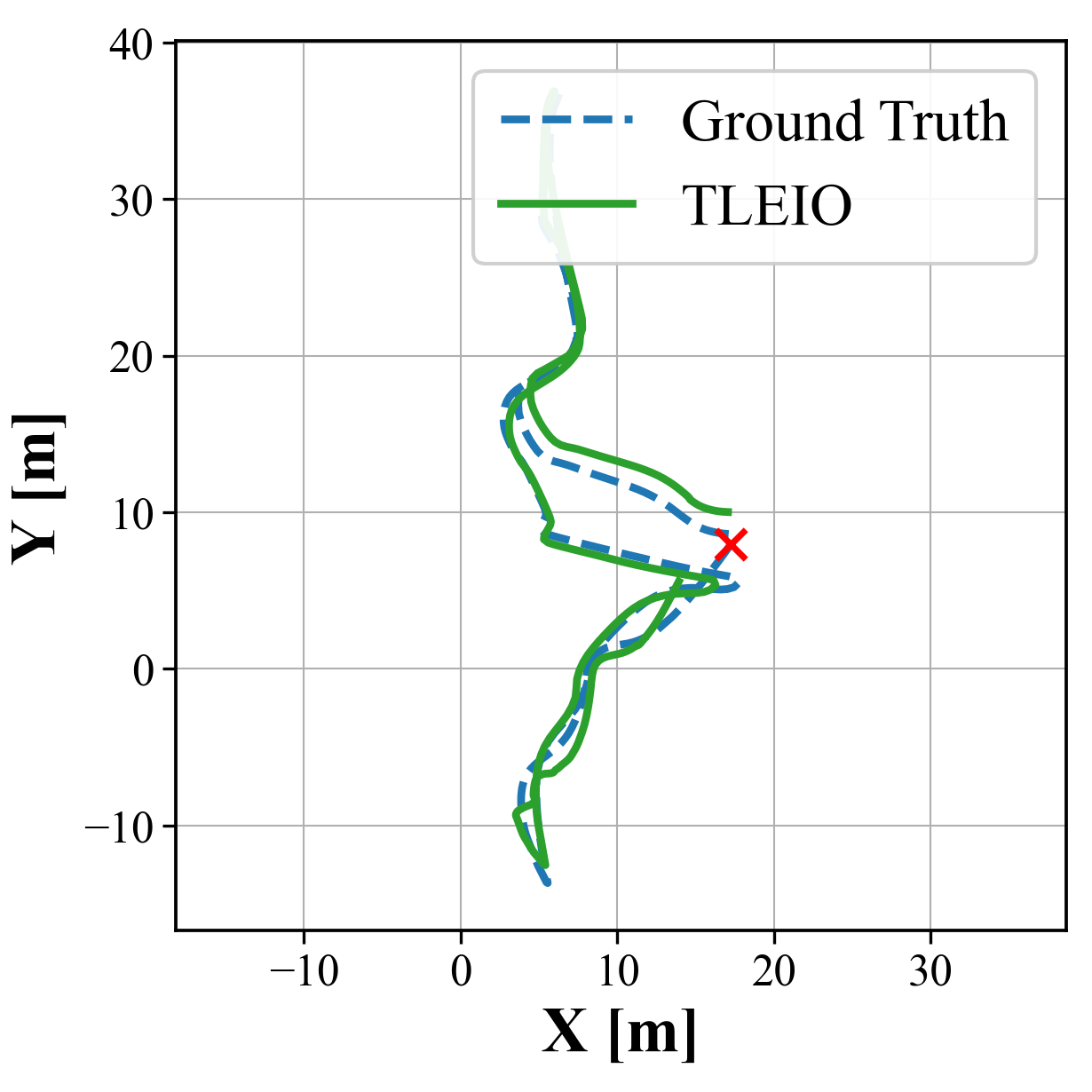}
\end{minipage}

\smallskip
\parbox[c][1.5em][c]{\linewidth}{\centering\small\textbf{(b) ME06}}\par
\begin{minipage}[c]{.65\linewidth}
\centering
\includegraphics[width=\linewidth]{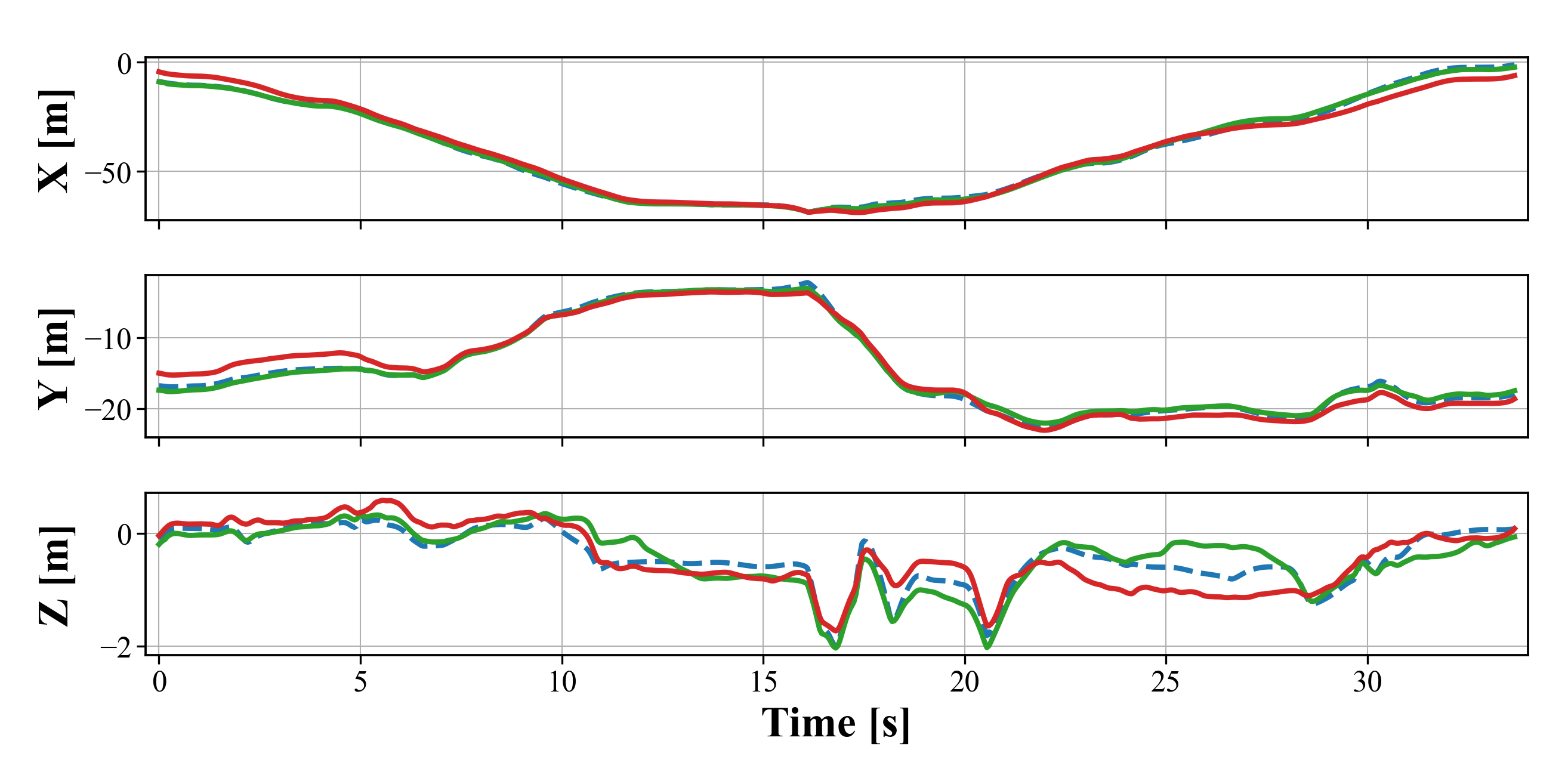}
\end{minipage}\hfill
\begin{minipage}[c]{.325\linewidth}
\centering
\includegraphics[width=\linewidth]{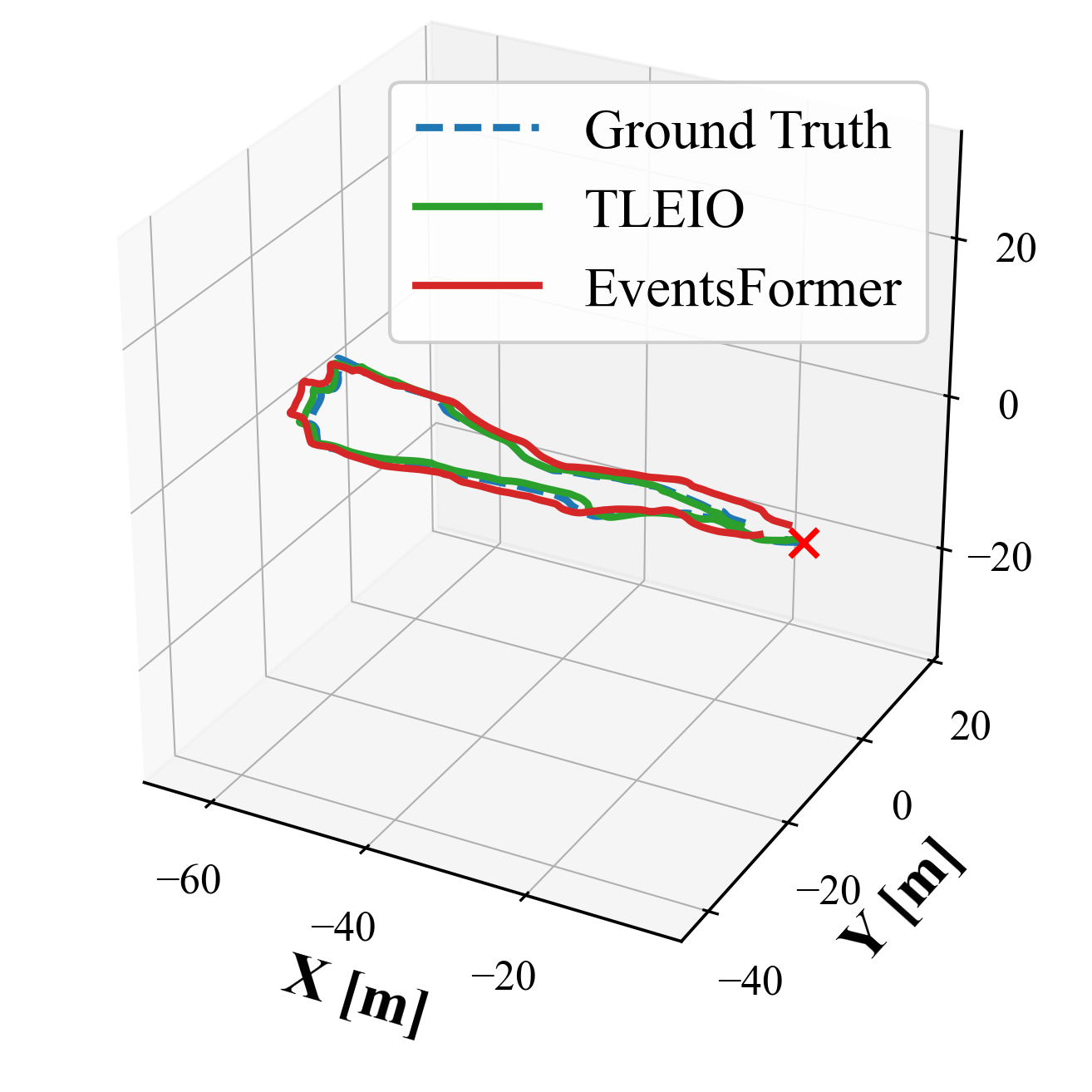}
\end{minipage}

\end{minipage}
\caption{TartanAir trajectories. (a,b) SE(3)-aligned position profiles and
spatial trajectories for MH02 and ME06; (b) includes standalone EventsFormer predictions, integrated using ground-truth orientations, 
to isolate the effect of filtering. }
\label{fig:tartanair_competition_results}
\end{figure}

The contribution of the event measurements is clearest on sequences where the
IMU-only baseline accumulates substantial error. On ME03, TLEIO reduces ATE
from $7.31$ to $1.37$\,m, while on MH03 it decreases from $3.72$ to
$0.22$\,m, corresponding to reductions of $81.3\%$ and $94.1\%$,
respectively. Overall, the lower error on 12 of 16 sequences
shows that the average gain is not limited to a few favorable cases and that
the learned event translation generally provides an effective correction to
inertial drift. Figure~\ref{fig:tartanair_competition_results} compares TLEIO
and standalone EventsFormer trajectories with the ground truth, while
Sec.~\ref{sec:runtime} reports the computational cost of the complete
pipelines.

\begin{table}[!htbp]
\centering
\caption{TartanAir competition split. ATE [m]; lower is better.
Published
baselines~\cite{Mahlknecht_2022_RAL,Pellerito_2024_IROS,Liu_2020_RAL,
Campos_2021_TRO,Schonberger_2016_CVPR,Teed_2021_DROID,
NEURIPS2023_7ac484b0} are reproduced from the RAMP-VO benchmark. F, E, and I
denote frames, events, and IMU; bold marks the lowest error in each row.}
\label{tab:tartanair_competition}
\footnotesize
\setlength{\tabcolsep}{2.2pt}
\renewcommand{\arraystretch}{1.00}
\begin{tabular*}{\textwidth}{@{\extracolsep{\fill}}lrrrrrrrr@{}}
\toprule
Sequence
& \multicolumn{1}{c}{\rotatebox{90}{EKLT-VIO}}
& \multicolumn{1}{c}{\rotatebox{90}{ORB-SLAM3}}
& \multicolumn{1}{c}{\rotatebox{90}{COLMAP}}
& \multicolumn{1}{c}{\rotatebox{90}{DROID-VO}}
& \multicolumn{1}{c}{\rotatebox{90}{DPVO}}
& \multicolumn{1}{c}{\rotatebox{90}{RAMP-VO}}
& \multicolumn{1}{c}{\rotatebox{90}{TLIO}}
& \multicolumn{1}{c}{\rotatebox{90}{\shortstack[c]{\textbf{TLEIO} \\ \textbf{(Ours)}}}} \\
\midrule
Input & \multicolumn{1}{c}{E+F+I} & \multicolumn{1}{c}{F} & \multicolumn{1}{c}{F}
& \multicolumn{1}{c}{F} & \multicolumn{1}{c}{F}
& \multicolumn{1}{c}{E+F} & \multicolumn{1}{c}{I} & \multicolumn{1}{c}{E+I} \\
\midrule
ME00 & 12.80 & 13.61 & 15.20 & 0.22 & \textbf{0.16} & 0.20 & 1.45 & 2.13 \\
ME01 & 9.10 & 16.86 & 5.58 & 0.15 & 0.11 & \textbf{0.04} & 2.30 & 1.20 \\
ME02 & 16.23 & 20.57 & 10.86 & 0.24 & 0.11 & \textbf{0.10} & 1.42 & 1.45 \\
ME03 & 18.08 & 16.00 & 3.93 & 1.27 & 0.66 & \textbf{0.46} & 7.31 & 1.37 \\
ME04 & 16.96 & 22.27 & 2.62 & 1.04 & 0.31 & \textbf{0.16} & 1.14 & 1.37 \\
ME05 & 6.84 & 9.28 & 14.78 & 0.14 & 0.14 & \textbf{0.13} & 1.85 & 1.19 \\
ME06 & 21.44 & 21.61 & 7.00 & 1.32 & 0.30 & \textbf{0.12} & 2.47 & 0.48 \\
ME07 & 9.06 & 7.74 & 18.47 & 0.77 & 0.13 & \textbf{0.12} & 1.37 & 0.72 \\
MH00 & 14.68 & 15.44 & 12.26 & 0.32 & \textbf{0.21} & 0.36 & 2.39 & 2.65 \\
MH01 & 4.35 & 2.92 & 13.45 & 0.13 & \textbf{0.04} & 0.06 & 0.94 & 0.14 \\
MH02 & 10.20 & 13.51 & 13.45 & 0.08 & \textbf{0.04} & \textbf{0.04} & 1.61 & 1.16 \\
MH03 & 2.13 & 8.18 & 20.95 & 0.09 & 0.08 & \textbf{0.04} & 3.72 & 0.22 \\
MH04 & 2.72 & 2.59 & 24.97 & 1.52 & 0.58 & \textbf{0.41} & 1.87 & 0.60 \\
MH05 & 4.48 & 21.91 & 16.79 & 0.69 & \textbf{0.17} & 0.25 & 1.35 & 0.48 \\
MH06 & 6.95 & 11.70 & 7.01 & 0.39 & \textbf{0.11} & \textbf{0.11} & 1.25 & 1.15 \\
MH07 & 19.72 & 25.88 & 7.97 & 0.97 & 0.15 & \textbf{0.07} & 1.66 & 1.44 \\
\midrule
Average & 10.98 & 14.38 & 12.50 & 0.58 & 0.21 & \textbf{0.17} & 2.26 & 1.10 \\
\bottomrule
\end{tabular*}
\end{table}

\begin{figure}[!htbp]
\centering

\makebox[\textwidth][c]{%
\begin{minipage}[t]{0.347\textwidth}
\vspace{0pt}
\centering
\parbox[c][1.5em][c]{\linewidth}{\centering\small\textbf{(a) Network}}
\includegraphics[width=\linewidth]{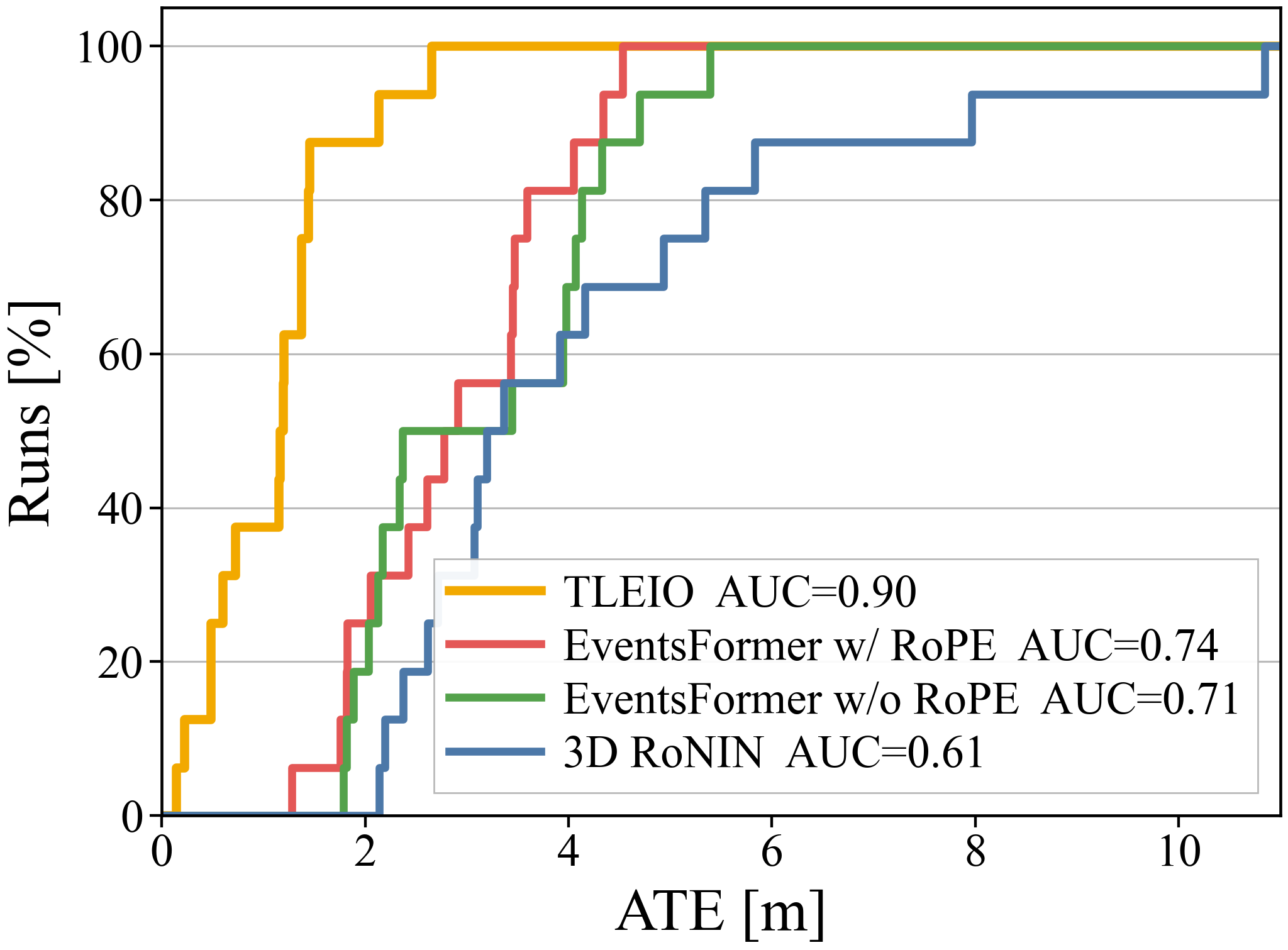}
\end{minipage}
\hspace{0.008\textwidth}
\begin{minipage}[t]{0.347\textwidth}
\vspace{0pt}
\centering
\parbox[c][1.5em][c]{\linewidth}{\centering\small\textbf{(b) Covariance}}
\includegraphics[width=\linewidth]{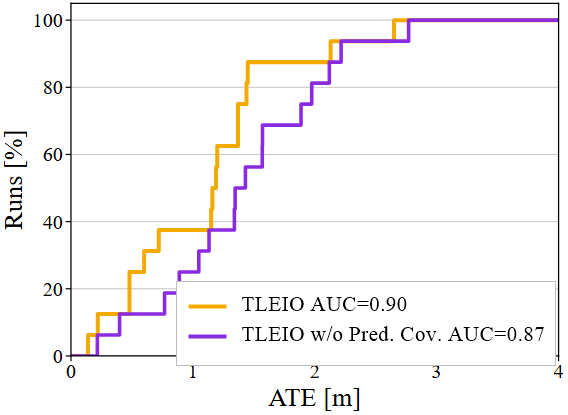}
\end{minipage}
\hspace{0.008\textwidth}
\begin{minipage}[t]{0.347\textwidth}
\vspace{0pt}
\centering
\parbox[c][1.5em][c]{\linewidth}{\centering\small\textbf{(c) Uncertainty}}
\includegraphics[width=\linewidth]{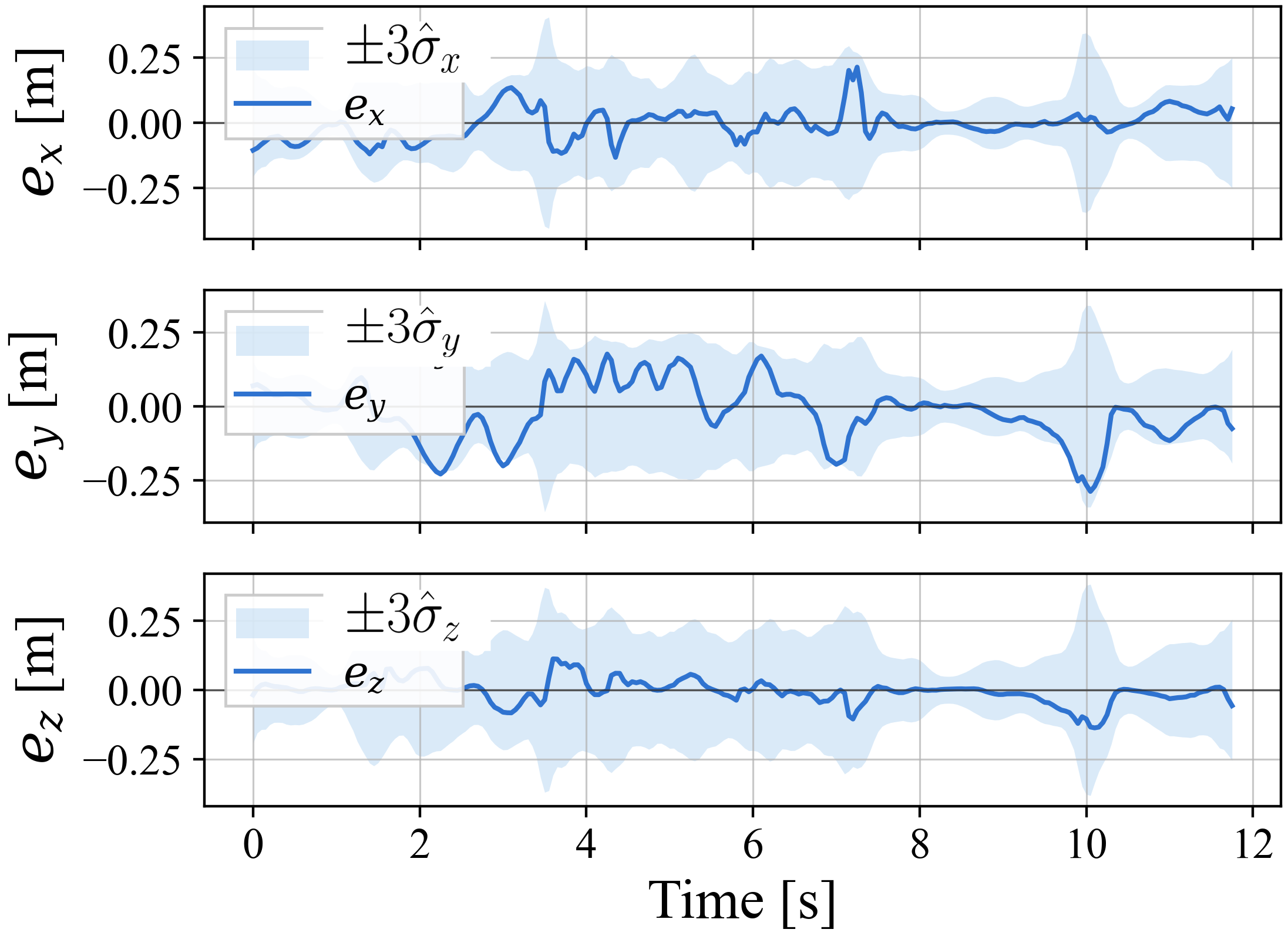}
\end{minipage}%
}

\vspace{0.15em}
\caption{TartanAir ablations.
(a) ATE distributions for the network and fusion variants.
(b) Learned versus fixed measurement covariance.
(c) Signed translation error and predicted
$\pm3\hat{\sigma}$ interval on MH01.}
\label{fig:tartanair_ablation_cdf}

\end{figure}

Figure~\ref{fig:tartanair_ablation_cdf} isolates the contribution of the learned
front-end, filter, positional encoding, and covariance prediction. In panel (a), we use the Area Under the Curve (AUC) metric to compare the standalone EventsFormer against the network-only 3D RoNIN model
used by TLIO~\cite{Liu_2020_RAL} and the whole TLEIO pipeline. 

Our pipeline reaches an AUC of $0.90$, compared with $0.74$ for
standalone EventsFormer with RoPE2D and $0.61$ for 3D RoNIN. This gap highlights the value of recursive
fusion, while the drop from $0.74$ to $0.71$ without RoPE2D confirms the benefit
of spatial positional encoding.

Panel (b) shows that learned covariance improves
ATE by up to $30\%$ over a fixed covariance, optimized for each sequence. The signed errors and
predicted $\pm3\hat{\sigma}$ intervals in panel (c) further illustrate how the
network adapts its uncertainty over time.

\subsection{Real-Data Adaptation on DAVIS240c}
\label{sec:davis240c_results}

We next test transfer to the real $240\times180$ DAVIS240c sensor, whose
handheld 6DoF motion, noise, and event statistics differ markedly from the
synthetic trajectories used for training. Following the published DEIO
protocol, we use $\mathrm{SE}(3)$ alignment over the first five seconds
~\cite{Guan_2025_ICCVW,Mueggler_2017_IJRR}. Adaptation updates only the final
linear prediction head, $9{,}240$ of approximately $31$M parameters, for 15
epochs on \texttt{dynamic\_translation} and \texttt{dynamic\_6dof}; the
transformer backbone remains frozen, and both training sequences are excluded
from evaluation.

\begin{table}[!htbp]
\centering
\caption{DAVIS240c held-out sequences. MPE [\%]; lower is better. Published
baselines
~\cite{Zhu_2017_CVPR,Mahlknecht_2022_RAL,Vidal_2018_RAL,Guan_2025_ICCVW,
Klenk_2024_3DV,Alzugaray_2019_3DV,Lee_2023_RAL,Rebecq_2017_BMVC} are
reproduced from the DEIO benchmark. F, E, and I denote frames, events, and IMU;
bold marks the lowest error in each column.}
\label{tab:davis240c_selected_subset}
\footnotesize
\setlength{\tabcolsep}{2.2pt}
\renewcommand{\arraystretch}{1.00}
\begin{tabular*}{\textwidth}{@{\extracolsep{\fill}}lcrrrrrrr@{}}
\toprule
Method & Input & \shortstack{boxes\\trans.} & \shortstack{HDR\\boxes}
& \shortstack{boxes\\6DoF} & \shortstack{poster\\trans.}
& \shortstack{HDR\\poster} & \shortstack{poster\\6DoF} & Avg. \\
\midrule
EKLT-VIO & E+F+I & 0.48 & 0.46 & 0.84 & 0.35 & 0.65 & 0.35 & 0.52 \\
Lee et al. & E+F+I & 0.74 & 0.69 & 0.77 & 0.28 & 0.52 & 0.59 & 0.60 \\
Rebecq et al. & E+I & 0.57 & 0.92 & 0.69 & 0.89 & 0.59 & 0.82 & 0.75 \\
HASTE-VIO & E+I & 2.55 & 1.75 & 2.03 & 1.34 & 0.57 & 1.50 & 1.62 \\
Zhu et al. & E+I & 2.69 & 1.23 & 3.61 & 0.94 & 2.63 & 3.56 & 2.44 \\
Ultimate-SLAM & E+I & 0.76 & 0.67 & 0.44 & 0.15 & 0.49 & 0.30 & 0.47 \\
DEIO & E+I & 0.07 & 0.09 & \textbf{0.05} & \textbf{0.04} & \textbf{0.06} & \textbf{0.08} & \textbf{0.07} \\
DEVO & E & \textbf{0.06} & \textbf{0.06} & 0.71 & 0.06 & 0.14 & 0.44 & 0.25 \\
\midrule
\textbf{TLEIO (Ours)} & E+I & 0.31 & 0.55 & 0.49 & 0.43 & 0.50 & 0.43 & 0.45 \\
\bottomrule
\end{tabular*}
\end{table}

\begin{figure}[t!]
    \centering

    \includegraphics[width=0.86\linewidth]
    {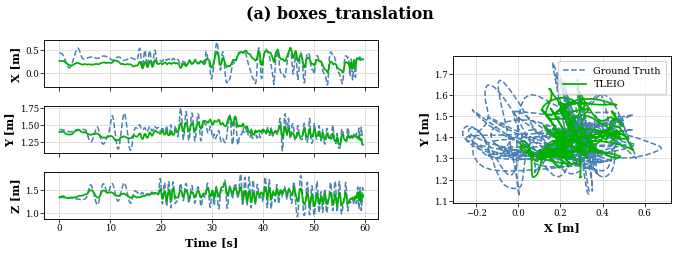}

    \vspace{1mm}

    \includegraphics[width=0.86\linewidth]
    {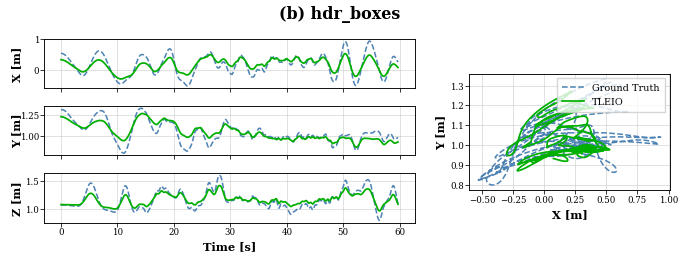}

    \vspace{1mm}

    \caption{
        Representative qualitative results on held-out DAVIS240c sequences.
        TLEIO trajectory estimates are compared with the corresponding
        reference trajectories on (a) \texttt{boxes\_translation} and
        (b) \texttt{hdr\_boxes}.
    }
    \label{fig:davis240c_qualitative}
\end{figure}

As Table~\ref{tab:davis240c_selected_subset} shows, TLEIO achieves an average
MPE of $0.45\%$, ranking third among nine methods and outperforming six of
the eight baselines, including DEVO on both 6DoF sequences. Representative
qualitative results are shown in Fig.~\ref{fig:davis240c_qualitative}, where
the estimated TLEIO trajectories are compared with the corresponding reference
trajectories on two held-out DAVIS240c sequences.

The lower average errors of DEVO and DEIO are expected from their heavier
back-ends: both track event-patch correspondences across multiple views and
iteratively refine poses and depth, while DEIO also incorporates IMU
preintegration in a sliding-window factor graph. TLEIO instead predicts local
translation directly from each event window and fuses it in a fixed-size
SC-EKF, without explicit matching or iterative refinement. It therefore trades
some accuracy for the simpler and substantially faster inference path reported
in Sec.~\ref{sec:runtime}.

\subsection{Runtime and Computational Profile}
\label{sec:runtime}
We evaluate the end-to-end runtime of TLEIO, RAMP-VO, DEVO, and DEIO on three EDS recordings~\cite{Hidalgo2022cvpr}. We use EDS because it provides real event-camera data at the same spatial resolution used during training. The selected recordings cover both low-light and well-lit scenes with event count ranging from 120.1 to 340.8 million events. Each native pipeline processes the complete recordings on the same RTX 4000 Ada 20\,GB GPU and Threadripper 7960X CPU. After two warm-up runs, we perform five timed runs and report end-to-end wall-clock time as mean $\pm$ sample standard deviation.

\begin{figure}[!htbp]
  \centering
  \includegraphics[width=\textwidth]{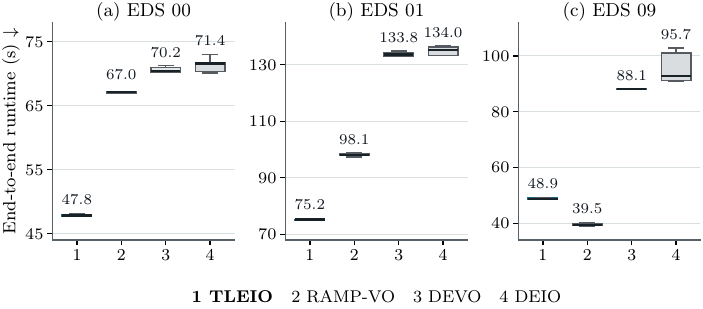}
  \caption{End-to-end runtime over five runs on EDS 00 (229.8M events),
  01 (340.8M), and 09 (120.1M). Boxes show the interquartile range, center
  lines indicate medians, and whiskers extend within $1.5\times$ the
  interquartile range. Labels report mean runtime in seconds; bold identifies
  the fastest method. Vertical ranges differ across panels.}
  \label{fig:runtime}
\end{figure}

TLEIO is the fastest method on sequences $00$ and $01$. On sequence $00$, it requires $47.847\pm0.196$\,s, reducing runtime by 28.6\% relative
to RAMP-VO, 31.9\% relative to DEVO, and 32.9\% relative to DEIO. On sequence $01$, TLEIO requires $75.205\pm0.260$\,s, with corresponding reductions of 23.3\%, 43.8\%, and 43.9\%.

On a mainly static sequence where the event count is lower (sequence $09$), our method is still constrained by its fixed-time voxel input. In this sequence RAMP-VO is fastest at $39.454\pm0.575$\,s, followed by TLEIO at $48.863\pm0.332$\,s. This result follows RAMP-VO's event-count-driven processing: on EDS, it forms a new input stack every $250{,}000$ events, so EDS $09$ produces approximately 480 network updates, compared with 919 for EDS $00$ and 1,363 for EDS $01$%
~\cite{Pellerito_2024_IROS}. Its lower runtime is therefore expected and explained by the smaller event count. TLEIO still requires 44.5\% less time than DEVO and 48.9\% less than DEIO, which also consume fixed-time voxels.

Summing the mean runtimes over all three recordings gives 171.915\,s for TLEIO, 204.555\,s for RAMP-VO, 292.130\,s for DEVO, and 301.051\,s for DEIO.
TLEIO therefore reduces the total runtime by 16.0\%, 41.2\%, and 42.9\%, respectively. Its inference path maps voxelized event windows directly to local displacement and uncertainty, avoiding feature extraction, keyframe graphs, and bundle adjustment. The fixed-size SC-EKF contains at most 45 error-state variables, uses a $12\times12$ innovation matrix, and accounts for only 5.4\% of TLEIO's measured runtime.

\section{Conclusions}
We presented TLEIO, a tight learned event-inertial odometry pipeline that combines EventsFormer, a Transformer-based front-end predicting short-window relative displacements from events, with a stochastic-cloning EKF driven by IMU propagation. To the best of our knowledge, TLEIO is the first event-inertial odometry framework treating events as the main source of learned motion measurements, while avoiding explicit event feature tracking, event-to-frame reconstruction, and pose-optimization backends at inference time.

Experiments on the TartanAir/TartanEvent competition split show competitive results against state-of-the-art odometry pipelines, outperforming IMU-only learned odometry and geometric event-inertial baselines. Despite the large domain shift from synthetic training data to real event-camera sequences, results on DAVIS240c further show that TLEIO adapts to different sensors with lightweight head-only fine-tuning, while preserving a compact learned front-end/filter interface. Overall, TLEIO provides a more efficient end-to-end alternative to optimization -based event odometry, achieving lower aggregate runtime than the evaluated optimization-based pipelines. These results support the central premise of this work: useful motion constraints can be learned directly from event streams in a single forward pass and effectively fused with inertial propagation to recover full 6DoF motion.

\bibliographystyle{splncs04}
\bibliography{main}

\begin{thebibliography}{10}
\providecommand{\url}[1]{\texttt{#1}}
\providecommand{\urlprefix}{URL }
\providecommand{\doi}[1]{https://doi.org/#1}

\bibitem{Gallego20pami}
Gallego, G., Delbruck, T., Orchard, G., Bartolozzi, C., Taba, B., Censi, A., Leutenegger, S., Davison, A., Conradt, J., Daniilidis, K., Scaramuzza, D.: Event-based vision: A survey. IEEE TPAMI  (2020). \doi{10.1109/TPAMI.2020.3008413}

\bibitem{Zhu_2017_CVPR}
Zihao~Zhu, A., Atanasov, N., Daniilidis, K.: Event-based visual inertial odometry. In: Proceedings of the IEEE Conference on Computer Vision and Pattern Recognition (CVPR) (July 2017)

\bibitem{Mahlknecht_2022_RAL}
Mahlknecht, F., Gehrig, D., Nash, J., Rockenbauer, F.M., Morrell, B., Delaune, J., Scaramuzza, D.: Exploring event camera-based odometry for planetary robots. IEEE Robotics and Automation Letters  \textbf{7}(4),  8651--8658 (2022). \doi{10.1109/LRA.2022.3187826}

\bibitem{Vidal_2018_RAL}
Vidal, A.R., Rebecq, H., Horstschaefer, T., Scaramuzza, D.: Ultimate {SLAM}? combining events, images, and {IMU} for robust visual {SLAM} in {HDR} and high-speed scenarios. IEEE Robotics and Automation Letters  \textbf{3}(2),  994--1001 (2018). \doi{10.1109/LRA.2018.2793357}

\bibitem{Lichtsteiner08ssc}
Lichtsteiner, P., Posch, C., Delbruck, T.: A 128$\times$128 120 db 15 $\mu$s latency asynchronous temporal contrast vision sensor. IEEE Journal of Solid-State Circuits  \textbf{43}(2),  566--576 (2008). \doi{10.1109/JSSC.2007.914337}

\bibitem{Weikersdorfer14icra}
Weikersdorfer, D., Adrian, D.B., Cremers, D., Conradt, J.: Event-based {3D} {SLAM} with a depth-augmented dynamic vision sensor. In: IEEE International Conference on Robotics and Automation (ICRA). pp. 359--364 (2014). \doi{10.1109/ICRA.2014.6906882}

\bibitem{Klenk_2024_3DV}
Klenk, S., Motzet, M., Koestler, L., Cremers, D.: Deep event visual odometry. In: Proceedings of the International Conference on 3D Vision (3DV). pp. 739--749 (2024). \doi{10.1109/3DV62453.2024.00036}

\bibitem{Guan_2025_ICCVW}
Guan, W., Lin, F., Chen, P., Lu, P.: {DEIO}: Deep event inertial odometry. In: Proceedings of the IEEE/CVF International Conference on Computer Vision (ICCV) Workshops. pp. 4665--4674 (2025)

\bibitem{Pellerito_2024_IROS}
Pellerito, R., Cannici, M., Gehrig, D., Belhadj, J., Dubois-Matra, O., Casasco, M., Scaramuzza, D.: Deep visual odometry with events and frames. In: IEEE/RSJ International Conference on Intelligent Robots and Systems (IROS) (2024)

\bibitem{Liu_2020_RAL}
Liu, W., Caruso, D., Ilg, E., Dong, J., Mourikis, A.I., Daniilidis, K., Kumar, V., Engel, J.: {TLIO}: Tight learned inertial odometry. IEEE Robotics and Automation Letters  \textbf{5}(4),  5653--5660 (2020). \doi{10.1109/LRA.2020.3007421}

\bibitem{4209642}
Mourikis, A.I., Roumeliotis, S.I.: A multi-state constraint kalman filter for vision-aided inertial navigation. In: Proceedings 2007 IEEE International Conference on Robotics and Automation. pp. 3565--3572 (2007). \doi{10.1109/ROBOT.2007.364024}

\bibitem{TimeSformer}
Bertasius, G., Wang, H., Torresani, L.: Is space-time attention all you need for video understanding? In: International Conference on Machine Learning (ICML). Proceedings of Machine Learning Research, vol.~139, pp. 813--824 (2021), \url{https://proceedings.mlr.press/v139/bertasius21a.html}

\bibitem{TSformerVO}
Françani, A.O., Maximo, M.R.O.A.: Transformer-based model for monocular visual odometry: A video understanding approach. IEEE Access  \textbf{13},  13959--13971 (2025). \doi{10.1109/access.2025.3531667}

\bibitem{zhu_voxels}
Zhu, A.Z., Yuan, L., Chaney, K., Daniilidis, K.: Unsupervised event-based learning of optical flow, depth, and egomotion. In: IEEE/CVF Conference on Computer Vision and Pattern Recognition (CVPR). pp. 989--997 (2019)

\bibitem{background_filter}
Guo, S., Wang, L., Chen, X., Zhang, L., Kang, Z., Xu, W.: {SeqXFilter}: A memory-efficient denoising filter for dynamic vision sensors. arXiv:2006.01687 (2020), \url{https://arxiv.org/abs/2006.01687}

\bibitem{heo2024ropevit}
Heo, B., Park, S., Han, D., Yun, S.: Rotary position embedding for vision transformer (2024), \url{https://arxiv.org/abs/2403.13298}

\bibitem{su2023roformer}
Su, J., Lu, Y., Pan, S., Murtadha, A., Wen, B., Liu, Y.: {RoFormer}: Enhanced transformer with rotary position embedding (2023), \url{https://arxiv.org/abs/2104.09864}

\bibitem{5462839}
Suh, Y.S.: Orientation estimation using a quaternion-based indirect kalman filter with adaptive estimation of external acceleration. IEEE Transactions on Instrumentation and Measurement  \textbf{59}(12),  3296--3305 (2010). \doi{10.1109/TIM.2010.2047157}

\bibitem{wang2020tartanairdatasetpushlimits}
Wang, W., Zhu, D., Wang, X., Hu, Y., Qiu, Y., Wang, C., Hu, Y., Kapoor, A., Scherer, S.: {TartanAir}: A dataset to push the limits of visual {SLAM} (2020), \url{https://arxiv.org/abs/2003.14338}

\bibitem{Mueggler_2017_IJRR}
Mueggler, E., Rebecq, H., Gallego, G., Delbruck, T., Scaramuzza, D.: The event-camera dataset and simulator: Event-based data for pose estimation, visual odometry, and {SLAM}. The International Journal of Robotics Research  \textbf{36}(2),  142--149 (2017). \doi{10.1177/0278364917691115}

\bibitem{Campos_2021_TRO}
Campos, C., Elvira, R., G{\'o}mez~Rodr{\'i}guez, J.J., Montiel, J.M.M., Tard{\'o}s, J.D.: {ORB-SLAM3}: An accurate open-source library for visual, visual-inertial, and multimap {SLAM}. IEEE Transactions on Robotics  \textbf{37}(6),  1874--1890 (2021). \doi{10.1109/TRO.2021.3075644}

\bibitem{Schonberger_2016_CVPR}
Sch{\"o}nberger, J.L., Frahm, J.M.: Structure-from-motion revisited. In: IEEE Conference on Computer Vision and Pattern Recognition (CVPR). pp. 4104--4113 (2016)

\bibitem{Teed_2021_DROID}
Teed, Z., Deng, J.: {DROID-SLAM}: Deep visual {SLAM} for monocular, stereo, and {RGB-D} cameras. In: Advances in Neural Information Processing Systems. vol.~34 (2021), \url{https://proceedings.neurips.cc/paper/2021/hash/89fcd07f20b6785b92134bd6c1d0fa42-Abstract.html}

\bibitem{NEURIPS2023_7ac484b0}
Teed, Z., Lipson, L., Deng, J.: Deep patch visual odometry. In: Oh, A., Naumann, T., Globerson, A., Saenko, K., Hardt, M., Levine, S. (eds.) Advances in Neural Information Processing Systems. vol.~36, pp. 39033--39051. Curran Associates, Inc. (2023), \url{https://proceedings.neurips.cc/paper_files/paper/2023/file/7ac484b0f1a1719ad5be9aa8c8455fbb-Paper-Conference.pdf}

\bibitem{Alzugaray_2019_3DV}
Alzugaray, I., Chli, M.: Asynchronous multi-hypothesis tracking of features with event cameras. In: Proceedings of the International Conference on 3D Vision (3DV). pp. 269--278 (2019). \doi{10.1109/3DV.2019.00038}

\bibitem{Lee_2023_RAL}
Lee, M.S., Jung, J.H., Kim, Y.J., Park, C.G.: Event-and frame-based visual-inertial odometry with adaptive filtering based on 8-dof warping uncertainty. IEEE Robotics and Automation Letters  \textbf{9}(2),  1003--1010 (2024). \doi{10.1109/LRA.2023.3339432}

\bibitem{Rebecq_2017_BMVC}
Rebecq, H., Horstschaefer, T., Scaramuzza, D.: Real-time visual-inertial odometry for event cameras using keyframe-based nonlinear optimization. In: Proceedings of the British Machine Vision Conference (BMVC) (2017). \doi{10.5244/C.31.16}

\bibitem{Hidalgo2022cvpr}
Hidalgo-Carri{\'o}, J., Gallego, G., Scaramuzza, D.: Event-aided direct sparse odometry. In: IEEE/CVF Conference on Computer Vision and Pattern Recognition (CVPR) (2022)

\end{thebibliography}

\end{document}